\documentclass[9pt,twocolumn,twoside]{osajnl}

\journal{ao} 

\setboolean{shortarticle}{false}

\title{The MODISSA testbed: a multi-purpose platform for the prototypical realization of vehicle-related applications using optical sensors}

\author[1,2]{Bj{\"o}rn Borgmann}
\author[1]{Volker Schatz}
\author[1]{Marcus Hammer}
\author[1,*]{Marcus Hebel}
\author[1]{Michael Arens}
\author[2]{Uwe Stilla}

\affil[1]{Fraunhofer IOSB, Ettlingen, Fraunhofer Institute of Optronics, System Technologies and Image Exploitation, Gutleuthausstr. 1, 76275 Ettlingen, Germany}
\affil[2]{Photogrammetry and Remote Sensing, Technical University of Munich (TUM), 80333 Munich, Germany}

\affil[*]{Corresponding author: marcus.hebel@iosb.fraunhofer.de}

\dates{\color{color_am1} \vspace{0.6em}\noindent \justifying \copyrightfont This is the authors' version of an article accepted for publication in Applied Optics, 9 May 2021. For the final edited and published version of record, see \href{https://doi.org/10.1364/AO.423599}{\color{color_am2}https://doi.org/10.1364/AO.423599}. Please cite as ``MODISSA: a multipurpose platform for the prototypical realization of vehicle-related applications using optical sensors'', Applied Optics 60 (22), F50-F65, 2021.

\vspace{0.4em} \noindent
\textcopyright\, 2021 Optical Society of America. One print or electronic copy may be made for personal use only. Systematic reproduction and distribution, duplication of any material in this paper for a fee or for commercial purposes, or modifications of the content of this paper are prohibited. \vspace{1em}}

\doi{\url{https://doi.org/10.1364/AO.423599}}

\begin{abstract}
We present the current state of development of the sensor-equipped car MODISSA, with which Fraunhofer~IOSB realizes a configurable experimental platform for hardware evaluation and software development in the contexts of mobile mapping and vehicle-related safety and protection. MODISSA is based on a van which has successively been equipped with a variety of optical sensors over the past few years, and which contains hardware for complete raw data acquisition, georeferencing, real-time data analysis, and immediate visualization on in-car displays. We demonstrate the capabilities of MODISSA by giving a deeper insight into experiments with its specific configuration in the scope of three different applications. Other research groups can benefit from these experiences when setting up their own mobile sensor system, especially regarding the selection of hardware and software, the knowledge of possible sources of error, and the handling of the acquired sensor data.
\end{abstract}

\setboolean{displaycopyright}{true}
\setboolean{displaydoi}{false}

\begin{document}

\maketitle


\renewcommand{\vec}[1]{\boldsymbol{#1}}

\gappto{\UrlBreaks}{\UrlOrds}

\makeatletter
\def\instring#1#2{TT\fi\begingroup
\edef\x{\endgroup\noexpand\in@{#1}{#2}}\x\ifin@}
\if\instring{.}{\thesubsection}\else
  \renewcommand\p@subsection{\thesection.}
  \renewcommand\p@subsubsection{\thesection.}
\fi
\makeatother


\section{Introduction}
\subsection{Motivation and context}
Optical sensor technologies and imaging are important areas of applied optics. Optical sensing provides a contactless method for acquiring information within a certain field-of-view. Besides the use of cameras for photography or television, such sensors can act as measuring devices for fast metric mapping of the physical world (photogrammetry). Furthermore, optical sensors form the basis for automatic perception (computer vision). The natural reference thereby is the combined function of eye and brain, which is to be reproduced or exceeded by technical perception systems, for example to allow for automation and control (robotics). Exceeding the natural example is achievable at different stages, for example in the computing speed and reaction time, the capacity for data storage or, above all, in the specific characteristics of the applied optical sensor technology. This may refer to the physical dimensions of the field-of-view, the spectral sensitivity of the detector, or the use of ``active sensors'' that measure the reflection of their self-generated signals in the scene. An example of the latter is scanning or imaging LiDAR (light detection and ranging), which uses laser pulses to obtain a pointwise three-dimensional representation of the environment.

There are many examples for the stationary use of optical sensors and for the automatic interpretation of the associated sensor data, e.g. sensors in industrial applications or surveillance cameras. In the sense of the preface and in analogy to human perception, however, the \textit{mobile} sensor operation is of special interest, since it is the only way to survey a larger area (example: the mars rover ``Perseverance''). Furthermore, a mobile sensor platform with real-time data interpretation enables self-related control of its movements by a computer (e.g., autonomous driving). Research in this field is currently being pursued with great effort. The driving force can primarily be seen in economic interests, e.g. of data services and automotive companies, but there is also politically driven support for these future technologies. In addition, the military sector plays a role.

\begin{figure*}[htbp]
\centering
\includegraphics[width=0.85\textwidth,viewport=18 147 771 481,clip=true]{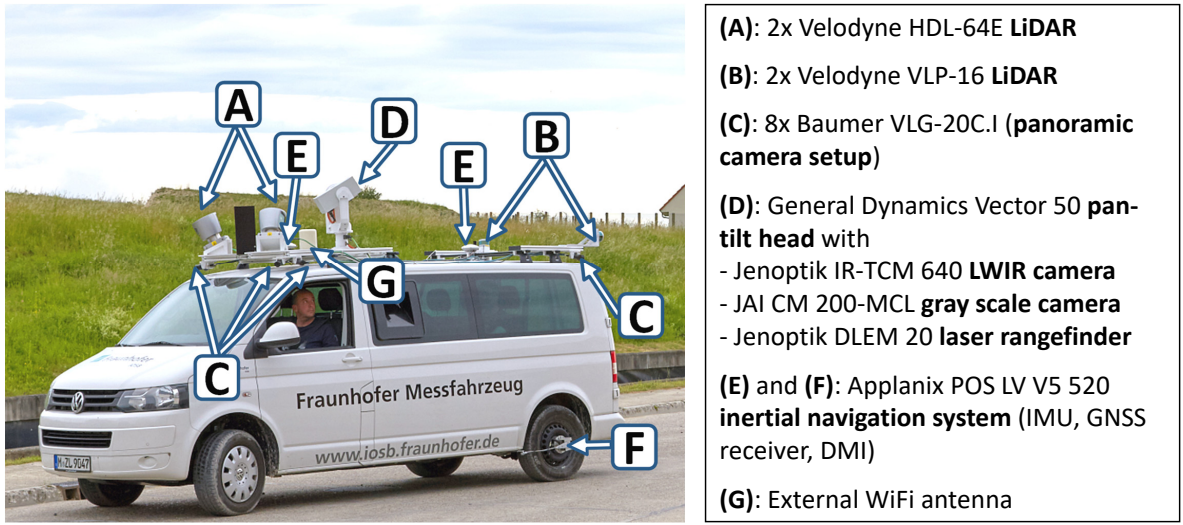}
\caption[MODISSA at a glance]{MODISSA at a glance: its appearance is characterized by a pan-tilt head, components of the inertial navigation system (INS), and several optical sensors.}
\label{fig:modissafig01}
\end{figure*}
Everyone can conduct research on these topics using publicly available datasets, without having to afford an own sensor-equipped measurement and testing platform. Such datasets are mainly provided by research groups in academia and by automotive companies to serve as standard benchmarks for performance comparisons, e.g., KITTI or nuScenes (see Section~\ref{sec:related}). However, datasets of this type can partially limit the intended research, because the selection of sensors was not under one's own control. Moreover, pre-recorded datasets cannot form the basis to study real-time situations in which immediate data interpretation affects the settings or the subsequent trajectory of the sensors. Such studies require an accessible research vehicle that provides test and analysis functionalities for a wide range of sensors and real-world operating conditions, which is why many groups have set up and now use such vehicles.
\subsection{Contribution of this paper}
With ``MODISSA'' as shown in Figure~\ref{fig:modissafig01}, also Fraunhofer~IOSB implements a configurable research vehicle that serves as an experimental platform for hardware evaluation and software development in the contexts of mobile mapping and vehicle-related safety and protection. Like a few companies and other research groups, we provide some publicly available test datasets. One particular example are mobile LiDAR scanning (MLS) data acquired from road traffic in an urban area~\cite{zhu2020}. However, pre-recorded benchmark datasets of this type have their limitations, as just described, and furthermore, such datasets are meanwhile available in sufficient quantity. In this paper, we instead describe our empirical findings in setting up and operating the MODISSA testbed over the past few years, and thus help others who intend to build their own version of a similar mobile sensor system. We provide a deeper insight into experiments in the context of three different applications, demonstrating the multi-purpose usability of our mobile sensor platform, and we discuss the pros and cons of the specific sensor selection and configuration we encountered in these applications.

After a brief review of related work in Section~\ref{sec:related}, we describe our experiences in setting up and operating MODISSA in Section~\ref{sec:setup}. Three of our current research applications that benefit from MODISSA as a working mobile sensor platform are described in Section~\ref{sec:applications}, with a summary of our respective lessons learned at the end of each subsection. Finally, Section~\ref{sec:conclusions} contains an overall discussion and some concluding remarks.
\section{Related work}
\label{sec:related}
In the context of the MODISSA experimental system presented here and examples of research conducted with it, three different categories of related work can be mentioned:
{\renewcommand{\theenumi}{(\arabic{enumi})}
\renewcommand{\labelenumi}{\theenumi}
\begin{enumerate}
	\item Work primarily concerned with the layout of experimental systems for mobile data acquisition or autonomous operation.
	\label{itm:rw1}
	\item Publications of datasets in this context, that are made available to the research community and are intended to serve as benchmarks for specific research tasks.
	\label{itm:rw2}
	\item Papers on research tasks and applications that require the use of a mobile data acquisition system or vehicle-mounted optical sensors.
	\label{itm:rw3}
\end{enumerate}}
From the user's perspective, category~\ref{itm:rw1} sets the stage for category~\ref{itm:rw2}, which in turn forms the basis for the actually relevant research in category~\ref{itm:rw3}. The latter covers applications from which entire scientific branches have evolved. To name just a few: situation awareness in road traffic, protection and driver assistance functions for vehicles, or simultaneous localization and mapping (SLAM). A comprehensive literature review on all possible applications would exceed the scope of this paper. We therefore limit this section to a selection of related work, mainly concerning categories~\ref{itm:rw1} and~\ref{itm:rw2}. Since this paper is intended to be part of an institutional focus issue that gives an insight into the work at Fraunhofer~IOSB, we address category~\ref{itm:rw3} by referring to a selection of our own previous publications throughout the paper, where appropriate.

Regarding~\ref{itm:rw1}: The capabilities for mobile data acquisition and computer-based interpretation of these data have greatly improved since the beginning of the century, when the use of digital sensors and powerful computers became realistically possible and affordable. This can also be seen by continuously increasing numbers of publications on this subject. In the above-mentioned category~\ref{itm:rw1} of related work, a distinction can again be made between those dealing in particular with autonomous driving and driver assistance~\ref{itm:rw1A} and those dealing with mobile systems for surveying~\ref{itm:rw1B}. With MODISSA, we can address real-time processing, but we can also record and georeference all sensor data. We thus aim at a multi-purpose applicability for research under both~\ref{itm:rw1A} and~\ref{itm:rw1B}.
{\renewcommand{\theenumi}{(\Alph{enumi})}
\renewcommand{\labelenumi}{\theenumi}
\begin{enumerate}
\item Work on autonomously operating cars received a significant boost from DARPA-sponsored competitions in the years 2004 to 2007, namely the ``DARPA Grand Challenge''~\cite{buehler2007} and the ``DARPA Urban Challenge''~\cite{buehler2009}. Both books~\cite{buehler2007} and~\cite{buehler2009} contain the original papers of all the important participants of these competitions, with comprehensive descriptions of their experimental systems. Research related to self-driving cars has continued rapidly since then and has increasingly spread from academia to the automotive industry. Some of the advances in methodology following the DARPA challenges are described in~\cite{pendleton2017}, and an overview of more recent developments in sensor technology for autonomous driving can be found in~\cite{campbell2018}. A related example is~\cite{zong2018}, where the authors describe and evaluate multiple sensor configurations and a software architecture used for their autonomous vehicle. The ``Formula Student Driverless'' (FSD) is a recent competition for autonomous racing cars that also encourages further development, and participants typically rely on a combination of cameras, INS, and LiDAR sensors for their very specific use case~\cite{kabzan2020}. With MODISSA itself, we do not aim at autonomous driving, but we can configure the sensors accordingly and thus work on algorithm development for this application. We additionally work on real-time detection of unexpected impacts on vehicles, for example by UAVs (cf. Subsection~\ref{subsec:uavs}).
\label{itm:rw1A}
\item Measurement vehicles for surveying and mobile mapping differ significantly from experimental systems for autonomous driving, since the quality of the data obtained is the main priority in this case. Consequently, sensors with high measurement accuracy and best coverage are used, together with inertial sensors for precise positioning of all measurements. The processing is usually performed separately from the data acquisition. An overview of sensor platforms for remote sensing can be found in~\cite{toth2016}. Vehicle-based, multi-sensor measurement systems of research institutions are described in~\cite{reiterer2020} and~\cite{paparoditis2012}, to name just two examples. Prominent examples of commercial measurement vehicles are the ones used by data companies like Alphabet or Here to collect data for their geodata or map services. A recent version of Google's Street View vehicles is equipped with a multi-camera setup to take spherical images of the surroundings, with two additional cameras to capture details, and with two Velodyne VLP-16 LiDAR sensors mounted at a 45{\textdegree} angle to acquire 3D data along the road~\cite{amadeo2017}. Here's cars are also equipped with LiDAR sensors and cameras~\cite{delaney2014}, and same as MODISSA, they employ ROS as part of their software stack.
\label{itm:rw1B}
\end{enumerate}}
Regarding~\ref{itm:rw2}: Over the years, companies and research groups provided test datasets recorded with their experimental systems. This is primarily being done to push competitive research in a particular direction with certain benchmarks, and it is often related to the work already mentioned in~\ref{itm:rw1}. Prominent examples are the well-known KITTI dataset~\cite{geiger2013} and extensions to it, e.g. KITTI-360~\cite{xie2016} or SemanticKITTI~\cite{behley2019}. Meanwhile, there are enough test and benchmark datasets available for different objectives, such as iQumulus~\cite{iqumulus2015} recorded with~\cite{paparoditis2012}, DublinCity~\cite{dublincity2019}, Paris-Lille-3D~\cite{parislille2018}, nuScenes~\cite{nuscenes2020}, the Lyft Level 5 dataset~\cite{lyft2020}, the Waymo open dataset~\cite{waymo2020}, ApolloScape~\cite{apolloscape2020}, Argoverse~\cite{argoverse2019}, or A2D2~\cite{a2d22020}. We too provide an MLS benchmark dataset for semantic labeling, recorded with MODISSA, which has been annotated at the Technical University of Munich~\cite{zhu2020}. However, as mentioned previously, we believe that research using pre-recorded data is limited in some cases, and that a dedicated experimental system is needed for deeper investigations.
\section{MODISSA: realization of a mobile multi-sensor system}
\label{sec:setup}
\subsection{Sensor selection and installation}
\label{subsec:installation}
\begin{figure}[htbp]
\centering
\includegraphics[width=\columnwidth,viewport=2 1 1260 650,clip=true]{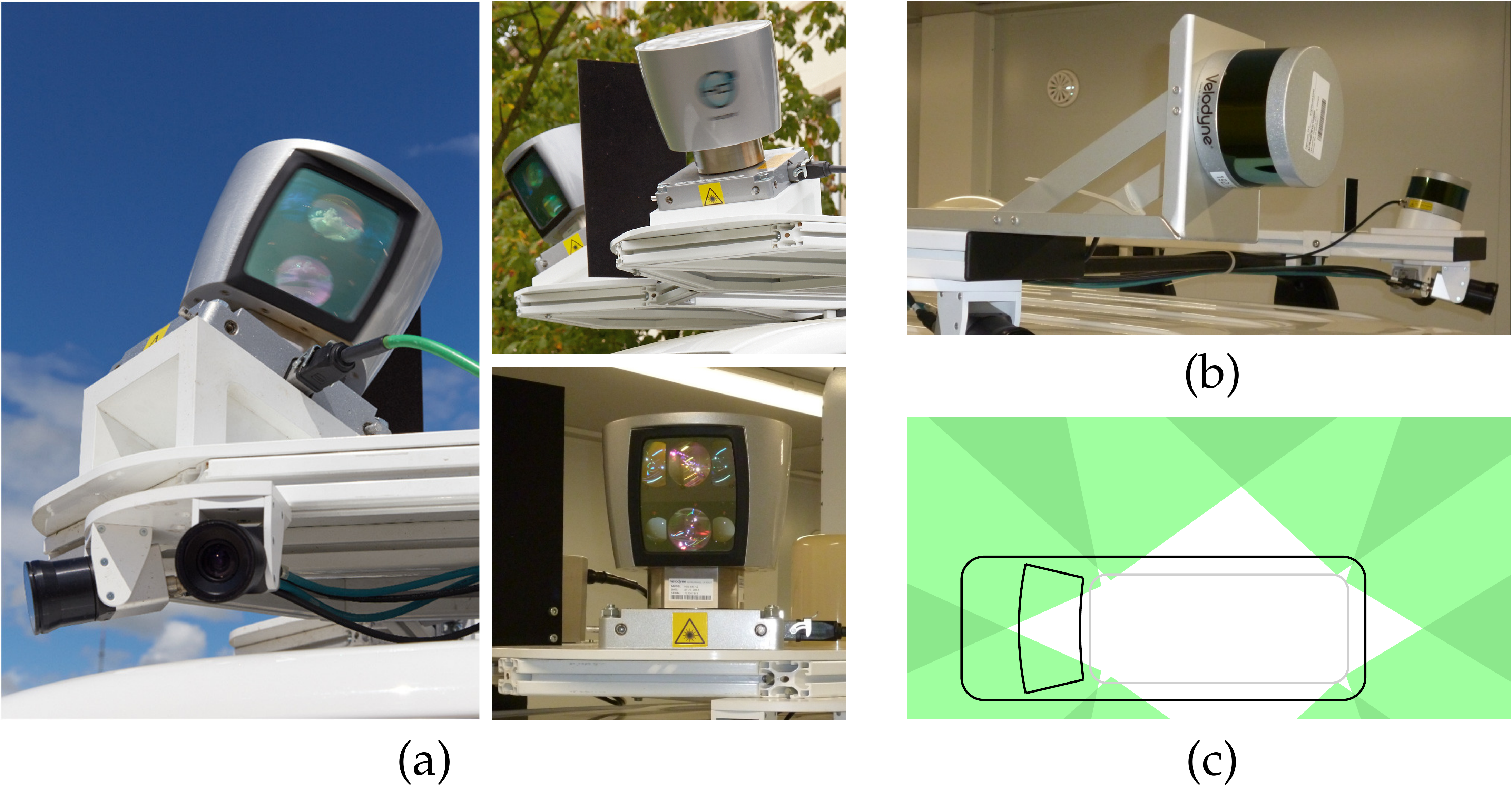}
\caption{Detailed views of the sensor installation: (a) mounting options for the front LiDAR sensors; 2 of the 8 panoramic cameras, (b) rear LiDAR sensor mounts, (c) bird's eye view of the fields-of-view of all panoramic cameras (vehicle front to the left).}
\label{fig:detailed-views}
\end{figure}

This subsection presents the sensors that are currently installed on MODISSA and why they were selected. Since the intended applications also dictate the most reasonable mounting options, we describe these along with the selection criteria. This paper, however, is not meant to be an advertisement for the named sensors. Both MODISSA's sensor configuration and its computing resources are being upgraded periodically, so we have generally avoided fully integrated solutions in favor of individual sensors we integrate ourselves. This is facilitated by the mechanical workshop of Fraunhofer~IOSB and by know-how in electronics and integrated circuit development in our department, which is the Department \textit{Object Recognition} (OBJ). The current sensor configuration is shown in Figure~\ref{fig:modissafig01}, and some detailed views can be seen in Figure~\ref{fig:detailed-views}.

\begin{itemize}
\item \textit{LiDAR sensors}: LiDAR is an established sensor technology for active 3D scene measurement. LiDAR sensors are available in a number of different types: as a range-imaging camera, flash LiDAR, or as a scanner. The achievable measurement ranges are affected by limitations on laser power to maintain eye safety at the specific laser wavelength. Currently, advances in LiDAR sensor technology with regard to longer ranges and better resolution can be noticed. Examples are Geiger mode, avalanche photodiode (APD), single-photon detectors, new detector materials, solid-state LiDAR, and digital LiDAR. The latter two terms refer to fewer mechanical and more highly integrated parts, promising continued cost reduction. 

Push-broom line scanners are typically the solution of choice for mobile mapping purposes, but they cannot provide a complete horizontal coverage of the vehicle's vicinity as required for real-time vehicle safety tasks. For MODISSA, we therefore opted for 360{\textdegree} 3D scanning LiDAR sensors, even though the accuracy and vertical coverage they can achieve for mapping tasks are typically lower than otherwise possible. First-generation devices, like those depicted in Figure~\ref{fig:detailed-views}a which we still use, will soon be replaced by devices with longer ranges and better resolution. Nevertheless, our application-related findings discussed in Section~\ref{sec:applications} remain valid and can be scaled to newer sensors.

MODISSA's front LiDAR sensors can be mounted on wedge-shaped bases in four azimuthal orientations. The inward tilted orientation shown on the left of Figure~\ref{fig:detailed-views}a has been used for detection of UAVs, and the outward tilt is suited for pedestrian detection and urban surveying, see Section~\ref{sec:applications}. The angle of the wegde is designed to compensate the downward angle of the lowest beam from the LiDAR sensor when it is looking towards the thick end of the wedge. The rear LiDAR sensors can also be mounted on a wedge base. A recent experimental change puts their rotation axis in the horizontal as shown in Figure~\ref{fig:detailed-views}b.
\item \textit{Panoramic camera setup}: An omnidirectional field-of-view around the vehicle is realized by eight identical video cameras attached in pairs to each corner of the vehicle roof. This setup was chosen after experiencing the drawbacks of an integrated spherical camera head, whose field-of-view was severely restricted by the vehicle roof and the other sensors. With the current setup, a parallax between individual cameras has to be tolerated, but full 360{\textdegree} coverage is achieved at less than 1.0~m distance from the vehicle on all sides. Figure~\ref{fig:detailed-views}c shows the overlapping fields-of-view. The cameras are 2~MPixel color cameras that output Bayer pattern data. That resolution is considered sufficient for our applications, and both the resolution and recording Bayer data keeps the amount of data low. The cameras are waterproof, and the lenses are covered by a waterproof tube attached to the cameras. This reduces the choice of cameras but makes it unnecessary to build cases for them. Further important requirements were a global shutter and external frame trigger capability, which allows synchronization with other sensors.
\item \textit{Inertial navigation system (INS)}: The INS is quite inconspicuous at first glance, but it has the greatest impact on consistent mobile data acquisition. In the commercial INS we use, a position computer (PCS) provides a position and orientation estimate based on input from a high-grade inertial measurement unit (IMU), a dual GNSS receiver, and a distance measurement indicator (DMI) attached to one of the rear wheels. The IMU is mounted on the same base plate on the vehicle roof as one of the front LiDAR sensors in order to minimize deviations between their movement. The INS allows GNSS-related errors to be reduced by including correction data obtained from a base station (real-time kinematic positioning, RTK). Next to inertial navigation, the PCS also provides a reference clock that is used for sensor synchronization, see Section~\ref{subsec:recording}.
\item \textit{Sensors on the pan-tilt head}:
A characteristic feature of MODISSA is the combination of its omnidirectional sensors with directional sensors on a pan-tilt head (PTH). The cameras on MODISSA's pan-tilt head are intended for dedicated purposes, namely 1. to resolve details of directly targeted objects, which is why optics corresponding to a narrower field-of-view were chosen, and 2. to capture objects in a multi-spectral manner, which is why a thermal resp. longwave infrared camera (LWIR) was placed alongside a visual camera (VIS). For the LWIR camera, an uncooled microbolometer camera was chosen, as this technology is lighter, less power-hungry and faster to power up than a photon detector IR camera. On the downside, microbolometer IR cameras always have a rolling shutter. The video camera has a snapshot shutter, and both cameras accept external frame triggers. Besides the two cameras, the PTH carries a laser distance meter, which allows to obtain the 3D location of an object sighted through the cameras.
\item \textit{Pan-tilt head}: In order not to limit our choice of cameras, we have opted for a general-purpose PTH instead of an integrated pan-tilt camera. Though the market for PTHs sold separately is small, we were able to obtain a type that combines a large payload (23~kg) with moderate weight (17~kg). It allows changing control parameters in the field, which is necessary for adaption to the payload. Its pan axis contains a slip ring with an allowance of user signals that we use to connect the cameras. This allows the PTH to rotate freely without any need to track the pan angle and unwrap external cables. For reliable operation of a camera with a Gigabit Ethernet (GigE) interface, we have replaced the slip ring with a variant designed for GigE and upgraded the user signal cabling. The PTH comes with an internal IMU for stabilized operation. Our control software uses a combination of prediction of the vehicle pose and the PTH stabilized mode to keep the sensors level.
\end{itemize}
The electrical power for the entire multi-sensor system and the computers is provided by large lithium-ion batteries and a power inverter. Up to 2000~W at 230~V AC can be provided continuously for more than five hours before they need to be recharged.

\begin{figure*}[htbp]
\centering
\includegraphics[width=0.85\textwidth,viewport=54 262 674 408,clip=true]{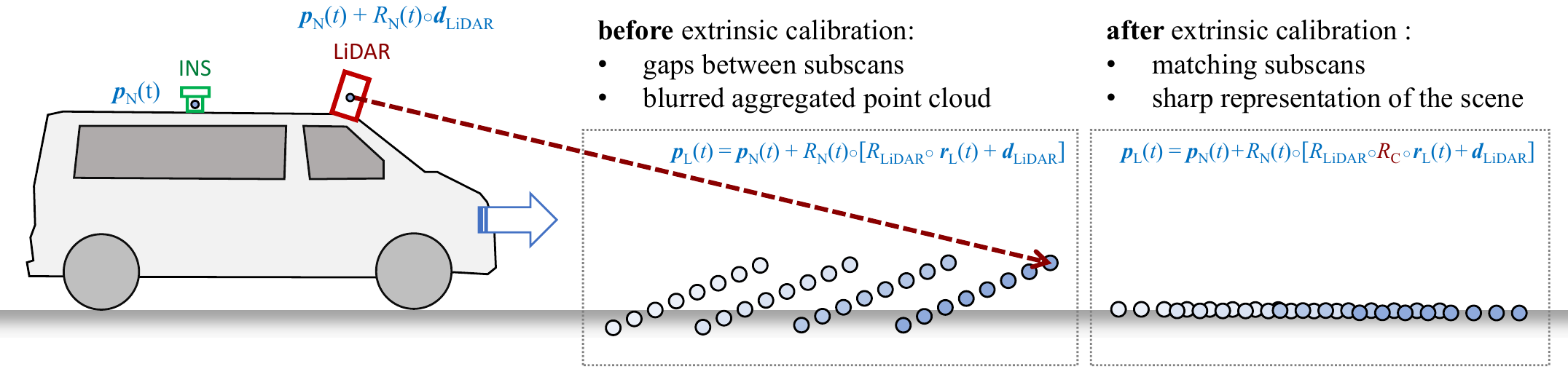}
\caption{Mobile LiDAR data acquisition with a newly assembled system typically results in a blurred aggregated point cloud. Only after extrinsic calibration, a sharp 3D representation of the scene can be acquired.}
\label{fig:extrinsic_cal}
\end{figure*}
Criteria for the orientation of the specific sensors have already been included in the previous list. For instance, the LiDAR sensors can be oriented to achieve best coverage along the road, in front of the vehicle, or towards the sky (see Figure~\ref{fig:detailed-views}), depending on the application being investigated.
The placement of the sensors on the vehicle is even more dependent on the mounting capabilities of the experimental platform. For example, in an experimental system, the sensors are more likely to be placed as a payload on a platform on the roof racks, e.g., to avoid losing the vehicle's road approval. The roof height above the road also has advantages in terms of the overview of the surrounding traffic. On the other hand, it is to be expected that other criteria will play a role towards series production, such as the drag coefficient of the vehicle and its aesthetic appearance. In a multi-sensor system, aspects of avoiding mutual occlusion or even mutual interference between the sensors also need to be taken into account. On MODISSA, shielding plates were placed between the LiDAR sensors specifically for this reason (see Figure~\ref{fig:detailed-views}a). Nevertheless, LiDAR crosstalk can still be observed~\cite{diehm2018}, but manufacturer's solutions for this problem are now being implemented for next-generation LiDAR sensors.
\subsection{Intrinsic calibration of the optical sensors}
\label{subsec:intrinsic}
A necessary preparatory step is the \textit{intrinsic calibration} of all optical sensors. In the case of cameras, the intrinsic parameters describe the optical center, the focal length, and the distortion of the lens that forms the image on the detector array. Standardized procedures and methods can be used to determine these parameters, such as can be found in the ``OpenCV'' program library (e.g., by using a ``checkerboard pattern''). In case of LiDAR sensors, intrinsic parameters may also include nonlinear corrections to the measured distances, the exact alignment of the individual laser rangefinders in a scanner head, or parameters describing the scanning process. Although the intrinsic parameters are usually determined and provided by the manufacturer, a custom intrinsic calibration procedure specific to the type of LiDAR sensor can further improve the data quality. Our group has also addressed this topic in the past~\cite{gordon2013}, but since it is a topic of its own and independent of MODISSA as a mobile data acquisition system, we do not go into further detail here.
\subsection{Extrinsic calibration and direct georeferencing}
\label{subsec:geo}
As already pointed out, a typical sensor-equipped car or mobile mapping system consists of several spatially separated parts. These are the optical sensors plus those of the INS, which typically comprises one or more GNSS receivers, wheel odometry (distance measurement indicator, DMI), and the inertial measurement unit (IMU) as its core element. Each optical sensor defines its own 3D coordinate frame, e.g. with the optical axis or the scan axis being one of the coordinate axes. The INS provides a navigation solution for its own well-defined coordinate frame, and measures its position and orientation in relation to the world coordinate system (e.g., ECEF). The vehicle's coordinate frame can be defined with a fixed transformation to that of the INS. Seen in this way, the vehicle is attached to the INS, not vice versa. The exact knowledge of the mutual placement and alignment of the INS's and each sensor's respective coordinate frame is of great importance for precise and consistent mobile data acquisition and interpretation, see Figure~\ref{fig:extrinsic_cal}. Using common terms, this refers to each sensor's \textit{lever arm} and \textit{boresight} orientation.

The procedure for geometric resp. \textit{extrinsic} system calibration outlined in this subsection concerns mobile data acquisition by LiDAR sensors (MLS). Similar methods are also being used for the cameras. We describe a condensed version of a calibration method that we previously published in~\cite{diehm2020}. In this overview paper on MODISSA, we focus on the essential boresight alignment, while omitting the more problematic but also rather unnecessary data-driven correction of the lever arm.

Let $\vec{d}_\text{LiDAR}$ denote the lever arm of the LiDAR sensor in question, and let $R_\text{LiDAR}$ be the estimate of its boresight orientation, such that $R^{-1}_\text{LiDAR}$ describes the alignment of the LiDAR sensor's coordinate frame relative to that of the INS. An initial estimate for $R_\text{LiDAR}$ usually exists by knowing the construction layout of the sensor carrier. As soon as position and orientation of one LiDAR sensor relative to the coordinate frame of the INS are determined, additional LiDAR sensors can be added to the transformation chain in terms of a LiDAR-to-LiDAR registration, e.g., by scanning reference surfaces and registering overlapping parts of the data. However, for at least one LiDAR sensor, a LiDAR-to-INS extrinsic calibration must be performed by mobile acquisition of LiDAR data involving the INS. The associated method described in this subsection is also applicable for in-field calibration, as it does not require a dedicated calibration setup.

Typically, opto-mechanical scanning provides a specific scan pattern, in which the distance $\rho_\text{L}(t)$ measured by the LiDAR sensor at time~$t$ is directed according to the scanning geometry. Let $\vec{r}_\text{L}(t)$ denote the 3D point measured by the LiDAR sensor in its own coordinate frame such that $\left\|\vec{r}_\text{L}(t)\right\|=\rho_\text{L}(t)$. Together with the navigational information $\vec{p}_\text{N}(t)$ and $R_\text{N}(t)$ describing current position and orientation in the world as measured by the INS, the LiDAR range measurements are directly georeferenced in the following way:
\begin{equation}
 \vec{p}_\text{L}(t) = \vec{p}_\text{N}(t) + R_\text{N}(t)\circ\left[R_\text{LiDAR}\circ\vec{r}_L(t) + \vec{d}_\text{LiDAR}\right]
\label{eq:dg}
\end{equation}
The aggregated points $\vec{p}_\text{L}(t)$ acquired within a time interval $\left[t_\text{1}, t_\text{2}\right]$ are usually called a \textit{LiDAR point cloud} or, more specifically, an \textit{MLS point cloud} in the case of mobile LiDAR scanning. Although the data acquisition is continuous, for convenience and by convention we usually split the data stream of georeferenced 3D points to a sequence of scans of 1/10~second duration, which in our system corresponds to single 360{\textdegree} 3D scans of the scanner head rotating at 10~Hz.

The accuracy of 3D point clouds obtained by direct georeferencing of LiDAR measurements is affected by several factors, reflecting the complexity of the sensor system. In addition to varying exactness of the navigational information sources (GNSS errors, IMU drift), several systematic effects can lead to reduced accuracy of point positioning. The lever arm of a LiDAR sensor can usually be determined with sufficient accuracy once the entire system is assembled, e.g., by simple measurement with a measuring tape. A boresight error has a much greater impact on the point positioning accuracy and typically occurs in magnitudes of some tenths of a degree on a newly assembled system. For example, a misalignment of 0.6{\textdegree} results in a displacement of 1.0~m for points at a distance of 100~meters. This subsection addresses the automatic correction of such boresight errors, assuming that all other influences were reduced to a minimum. In this sense, extrinsic calibration of the mobile LiDAR system means the automatic determination of a boresight correction $R_\text{C}$ to Eq.~\ref{eq:dg} in the following way:
\begin{equation}
 \vec{p}_\text{L}(t) = \vec{p}_\text{N}(t) + R_\text{N}(t)\circ\left[R_\text{LiDAR}\circ R_\text{C}\circ\vec{r}_L(t) + \vec{d}_\text{LiDAR}\right]
\end{equation}

To calibrate the system and find $R_\text{C}$ automatically, we assess and optimize the quality of the MLS point cloud $P_{t_\text{1},t_\text{2}}$ acquired in a certain time interval $\left[t_\text{1}, t_\text{2}\right]$. For a point $\vec{p}_\text{i}$, let $\left\{\vec{p}_\text{i}\right\}_\text{N}$ be the set of $\vec{p}_\text{i}$ and its $\text{N}$ nearest neighbors. The smallest eigenvalue $\lambda_{\text{i}1}$ found by principal component analysis (PCA) of this neighborhood quantifies the local scatter of 3D points at this position. We intend to minimize the average local scatter $S$ of the 3D points such that the resulting point cloud is a sharp 3D representation of the scene (cf.~Figure~\ref{fig:extrinsic_cal}). Hence, with the total number $n_p$ of points in the point cloud $P_{t_\text{1},t_\text{2}}$, we require 
\begin{equation}
 S = \frac{1}{n_p\cdot(\text{N}+1)}\sum_{\text{i}=1}^{n_p}\lambda_{\text{i}1} \hspace{1em} \rightarrow \text{min}
\end{equation}
A simple but easy to implement numerical way to find the minimum of $S$ is to perform a full or iterative grid search in the parameter space whose three dimensions are the Euler angles of $R_\text{C}$.

\begin{figure}[htbp]
\centering
\includegraphics[width=\columnwidth,viewport=62 148 682 476,clip=true]{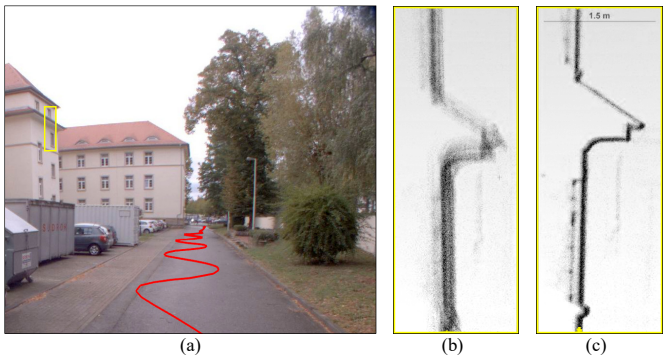}
\caption{(a) View of an appropriate scenario used for the calibration. (b) LiDAR points in the yellow section acquired with the uncalibrated system. (c) The same points after extrinsic calibration.}
\label{fig:extrinsic_cal_real}
\end{figure}
In analogy to the previously shown illustration, Figure~\ref{fig:extrinsic_cal_real} gives an impression of the calibration procedure at a real-world scenario. Figure~\ref{fig:extrinsic_cal_real}a shows the path driven to acquire the data used for the calibration (red), and an exemplary section of a building (yellow). Figures~\ref{fig:extrinsic_cal_real}b and~\ref{fig:extrinsic_cal_real}c show the LiDAR points in this section before and after the calibration procedure.

\textit{Results and lessons learned}:
We conducted several experiments to evaluate the suitability of different terrains as well as influences of the driving maneuver and other boundary conditions on the calibration process. Some results of these experiments are described in our previous paper~\cite{diehm2020}, and we have since been able to confirm and extend these findings. The following results can be noted: For an extrinsic calibration in the proposed manner, the LiDAR-to-INS extrinsic calibration works best when only short time intervals $\left[t_\text{1}, t_\text{2}\right]$ are considered, during which a stable GNSS-based positioning can be expected. In our calibration runs, these time intervals are therefore typically no longer than 20~seconds. Within the time interval, curvy driving maneuvers (e.g. zigzag, see Figure~\ref{fig:extrinsic_cal_real}a) should be performed in a terrain with enough surfaces with varying orientations. Among the terrain types typically available, urban areas are the most suitable. Here, we succeeded in determining the orientation angles of MODISSA's LiDAR sensors with an estimated residual error less than 0.1{\textdegree}, which corresponds to a point deviation of 17~cm at a distance of 100~m.
\subsection{Synchronized sensor data acquisition and recording for mobile surveying}
\label{subsec:recording}
When using MODISSA as an experimental mobile sensor platform for a particular application, we distinguish between its \textit{data recording system} and its \textit{real-time processing system}, in the same sense as the distinction~\ref{itm:rw1B} vs.~\ref{itm:rw1A} was made in Section~\ref{sec:related}. This distinction is not actually visible from the outside, because both systems share the same infrastructure of the installed sensors and their synchronization, and both systems can even run in parallel. In this subsection, we describe the implementation of synchronized sensor data acquisition and recording, which is based on the experiences gained with our previous airborne data acquisition platform~\cite{schatz2008}. The real-time processing system is described separately in Subsection~\ref{subsec:setup_realtime}.

\begin{figure}[htbp]
\centering
\includegraphics[width=\columnwidth,viewport=70 64 786 560,clip=true]{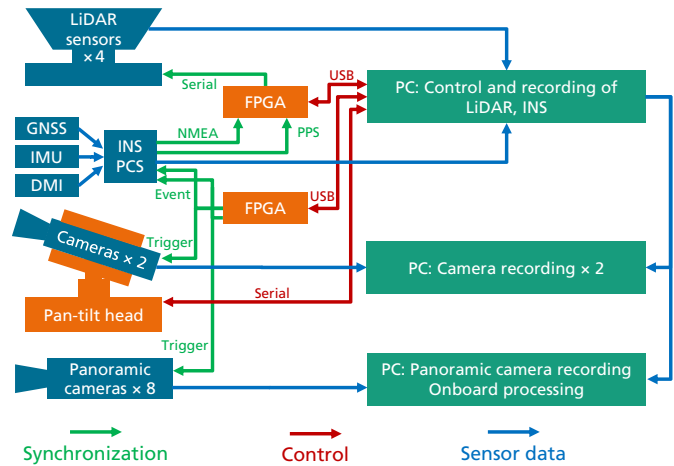}
\caption{Sensor synchronization, sensor control paths, and sensor data transmission in the data recording system.}
\label{fig:data_flow_overview}
\end{figure}
Figure~\ref{fig:data_flow_overview} gives an overview of the several data and control links in the recording system. The data stream of each sensor is recorded by one of a total of four computers. These computers are connected to each other in a network so that their sensor data can also be shared, e.g., in the context of real-time processing (see Subsection~\ref{subsec:setup_realtime}). One of the computers is responsible for the sensors with low data volume, i.e., the INS and the LiDAR sensors. It is also used for several control tasks, including control of the pan-tilt head, thus it also records the pan-tilt head's orientation data. Both the visual and thermal camera installed on the pan-tilt head are operated with a dedicated recording computer. The outputs of the eight cameras of the panoramic camera setup are recorded by the fourth computer, which is also the one to be used for the real-time processing.

An IOSB proprietary data format is used to stream and record all image sequences (SIS, the ``IOSB Image-Sequence File format''~\cite{scherer2012}. It is uncompressed to facilitate random access when working with recorded data. It supports supplemental metadata for each frame as well as for the entire recording.

When recording multiple data streams, time tagging of all data is required to enable time-synchronous data fusion in subsequent post-processing operations. The position computer (PCS) of the INS also contains the reference clock for the sensor system. It can transmit the position and time via several serial interfaces in the format of ``National Marine Electronics Association'' (NMEA) messages together with a synchronization pulse indicating the start of each second (PPS). This is the synchronization method used for the LiDAR sensors, which are designed to accept such messages. As shown in Figure~\ref{fig:data_flow_overview}, in our case a translation device is interposed between the PCS and the LiDAR sensors. It contains a small circuit implemented at Fraunhofer~IOSB in a field-programmable gate array (FPGA) that serves multiple purposes. The Velodyne HDL-64E requires a specific NMEA message and the PPS on the same signal input within a narrow time window relative to each other. The PCS has not always supported the required NMEA message type, transmits the PPS on a separate line and does not observe the required timing. The translation box delays the NMEA message as needed and retransmits it to all four LiDAR sensors. In addition, it allows to send commands it receives from the control computer to one of the Velodyne HDL-64E sensors. This has been used for experiments in synchronizing the rotation of the Velodyne HDL-64E sensors by continually adapting the rotation speed of one of them to avoid LiDAR crosstalk.

As a second synchronization method, the PCS provides so-called event inputs that can receive signal pulses. For every pulse detected, the PCS generates a timestamp and inserts it into the navigation data stream. This is used for the cameras. All cameras run in \textit{external frame trigger mode}, which means that each individual image they capture is triggered by a signal coming from outside the cameras. The trigger generator is again a custom circuit implemented in an FPGA. All trigger signals are duplicated and transmitted to both the cameras and an event input of the PCS. Separate signals are used for the LWIR camera and the visual camera on the pan-tilt head, as well as for the group of eight panoramic cameras. Therefore, each of these cameras (or group of cameras) can be operated at an individual frame rate.

There may be a small delay between the trigger pulse and the start of the exposure of a video camera. We have a way to measure it~\cite{schatz2017} and have found it to be of the order of microseconds. Knowledge of this delay and the exposure time allows to shift the timestamps to the middle of the exposure interval in offline processing. Alternatively, a feature in our custom trigger generator allows to shift the trigger times by a set amount and thereby compensate a constant offset. Microbolometer IR cameras such as the one on the PTH have a much more complex timing behavior. Due to their operating principle, there is no exposure interval and the pixel value is determined by the observation of the scene preceding the trigger time. We also have a method (as yet unpublished) to measure their time constant, which is of the order of milliseconds.

The recording computers receive a copy of part of the navigation data stream from the control computer. The timestamp and an interpolated position and orientation are embedded into the frame headers of the image data streams. Triggering is started with a slight delay after starting data acquisition so that recording cannot miss the first frame. After that, frames and timestamps are matched by their sequence numbers. Due to the use of non-real-time operating systems, it can happen that the two data streams temporarily drift apart so much that the camera stream would have to be delayed more than desired. In that case, the embedding skips some frames and continues when the navigation data stream has caught up. The current angles of the PTH are embedded in the same way, though there is no hardware support for properly synchronizing those values.

\subsection{Real-time data processing and sensor control}
\label{subsec:setup_realtime}
The research tasks of type~\ref{itm:rw1A} specified in Section~\ref{sec:related} require a capability for immediate data processing. To this end, the \textit{real-time processing system} was established on MODISSA, which is described in more detail in this subsection. In terms of the computers and software used, it is still designed to facilitate research activities. This implies that the focus in setting up the real-time processing system is on configurability and ease of operation, rather than on a final, technologically optimized system design. Constraints of compactness and energy efficiency were therefore not a primary concern. For the computer hardware, this means that a standard PC environment is used for the onboard processing, which has a high compatibility with existing C++ algorithm implementations from related research projects. 

The software environment for real-time processing on MODISSA is based on the ``Robot Operating System'' (ROS, \href{https://www.ros.org}{www.ros.org},~\cite{quigley2009}). Although ROS is originally intended for its use in robotics, it is also useful for multi-sensor systems in general. The software environment facilitates the communication between multiple software processes and provides an abstraction layer between the sensors, other hardware components, and the data processing. The abstraction layer allows changes to the sensor equipment without requiring extensive changes to the implementation of methods for data analysis. It provides access to sensor data in standardized formats, access to calibration information of respective sensors, the \textit{kinematic chain}, standard processing of sensor data, and easy-to-use interfaces for controlling actuators, such as the pan-tilt head.

\begin{figure}[htb]
	\begin{center}
		\includegraphics[width=\columnwidth]{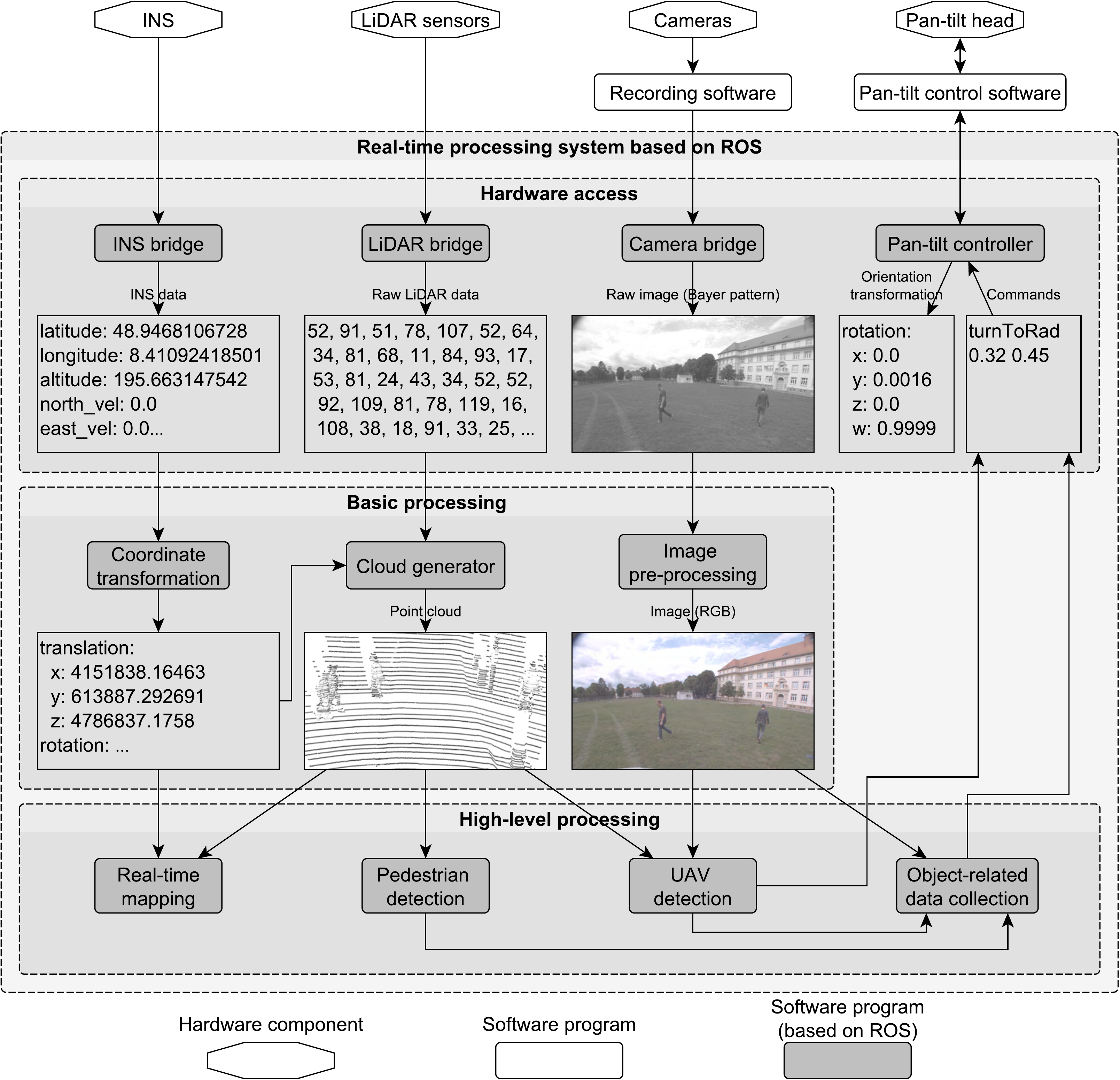}
		\caption{Schematic overview of the ROS-based system for real-time processing. Exemplary building blocks illustrate its modular applicability.}
		\label{fig:real-time_processing}
	\end{center}
\end{figure} 
Figure~\ref{fig:real-time_processing} shows a schematic overview of MODISSA's real-time processing system. The different program modules communicate with each other through the mechanisms of ROS. The system can be divided into three layers: The first layer communicates with the hardware, either directly or, in the case of the cameras and the pan-tilt head, through programs of the recording system (see Subsection~\ref{subsec:recording}). This layer is the only one that interacts with components outside the ROS-based environment. The second layer provides some standard pre-processing steps for the sensor data. For example, RGB images are generated from the raw Bayer pattern data of the panoramic cameras. In the case of the LiDAR sensors, this layer includes the real-time generation and direct georeferencing of 3D point clouds. The INS measurements are used to provide and update transformations between coordinate frames of the individual sensors and the global geocoordinate system. The third layer contains the software modules of the actual high-level processing, which are related to the specific research tasks and are developed individually for this purpose (see Section~\ref{sec:applications}).
\subsection{Data protection}
\label{subsec:anonym}
Data collection in road traffic within populated areas generates some sensitive data, e.g., image recordings of license plates and persons with time and location references. A factor that is often initially underestimated in this context is compliance with the data protection regulations of the respective country. In the European Union (EU), the processing of personal data is prohibited unless at least one of six different reasons legitimizes it, one of which being the ``legitimate interest'' of the processor (Art.~6 of the EU's General Data Protection Regulation, GDPR). In this case, the data processor must ensure that this interest is not overridden by the interests or fundamental rights and freedoms of the data subject. Therefore, a high level of data protection and data security is mandatory. In preparation of the initial activities with MODISSA, we obtained a lawyer's expertise and then defined a data protection concept according to which we now plan, conduct, and post-process our measurement runs. Nevertheless, this section cannot and should not be understood as a legal advice, we only give a brief insight into this data protection concept. It is based on three core principles:
\begin{itemize}
	\item \textit{Avoidance}: The acquisition and storage of sensitive data is avoided whenever possible. The reasons to record data are clearly justified and documented in advance in a measurement plan.
	\item \textit{Security}: The duration of data storage is minimized and access to these data is secured at all times by technical measures. Access to the mobile sensor platform and the data is limited to a necessary number of persons. Each measurement run on public ground is documented in a logbook.
	\item \textit{Privacy preserving}: For the subsequent scientific use that is described in the measurement plan, the collected data are promptly subjected to an automated anonymization process.
\end{itemize}

\begin{figure}[htbp]
\centering
\includegraphics[width=\columnwidth,viewport=1 238 751 403,clip=true]{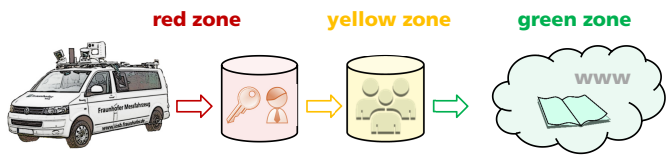}
\caption{Different zones for the recorded data: red, yellow, and green zone, each with a different balance between access restriction and anonymization.}
\label{fig:dataprotectzones}
\end{figure}
With this in mind, we have defined three different environments where the data are processed, see Figure~\ref{fig:dataprotectzones}. In each environment, appropriate measures enforce the three core principles:
\begin{itemize}
\item In the \textit{red zone} we handle temporarily stored raw data. It has the highest level of access restrictions, physically and by encryption. The only allowed access to it is to anonymize the data to transfer them into the yellow zone.
\item The \textit{yellow zone} contains algorithmically anonymized data for research purposes with limited access by people who need the data for their research.
\item In the \textit{green zone} we put selected and manually inspected anonymized data for print and online publications.
\end{itemize}

\begin{figure}[htbp]
\centering
\includegraphics[width=\columnwidth,viewport=47 227 663 425,clip=true]{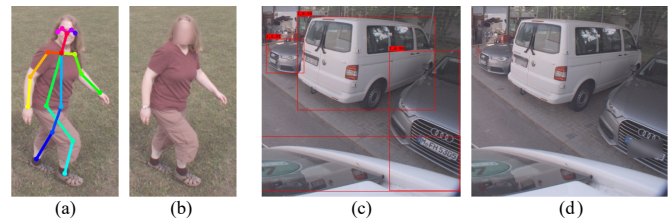}
\caption{(a) Person with facial region detected by OpenPose, (b) anonymization result. (c) Cars detected by YOLOv3, (d) license plates located and blurred. Results from~\cite{muench2019}.}
\label{fig:dataprotectexamples}
\end{figure}

The main element of our data protection concept is the automatic anonymization of personal data, which transfers the sensor data from the red zone to the yellow zone. It has been implemented in our department at Fraunhofer~IOSB using modern machine learning methods, see~\cite{muench2019}. The technical details can be summarized as follows: we use ``OpenPose'' to estimate keypoints of persons, locate their facial regions and then blur them (Figures~\ref{fig:dataprotectexamples}a and b). Furthermore, we use a two-step ``YOLOv3'' approach to detect and locate vehicles, and then to find and blur their license plates (Figures~\ref{fig:dataprotectexamples}c and d).

\textit{Results and lessons learned}:
A comprehensive performance evaluation of our anonymization methods can be found in~\cite{muench2019}. The values for precision and recall obtained in these evaluations indicate recognition performance on an almost human level, such that the final manual inspection to transfer sensor data to the green zone is reduced to a minimum. 

Since the sensor data are recorded to be used as the basis for subsequent scientific studies, partially blurred image data may no longer be usable depending on the specific research topic, e.g., for the detection of distracted pedestrians or facial emotions. The only option in such cases is to stage scenes on private ground.
 
Beyond compliance with the legal requirements, it may be good practice to inform the local authorities about the measurement campaigns in case they receive any inquiries or complaints. For the same reason, it can be helpful to carry information material about the measurement vehicle and the objective of the measurements, which can be passed on to interested citizens upon request.
\section{Selected applications, methods, and results}
\label{sec:applications}
Depending on the particular research context, dedicated measuring vehicles or sensor-equipped cars are often specifically designed to achieve the research goals for the application at hand. With MODISSA, we aim at a multi-purpose applicability with simultaneous adaptability, so that we can address both real-time applications and data recording for later offline processing. In this section, we give examples of both use cases. More specifically, these selected applications are: (A) mobile 3D mapping, (B) LiDAR-based person detection, and (C) multi-sensor UAV detection and tracking. These examples are part of current research projects at Fraunhofer~IOSB and were selected from a number of others, in which the MODISSA vehicle plays an essential role.
\subsection{Area-wide mobile 3D mapping}
The term \textit{mobile mapping} describes the process of acquiring and recording geospatial data using a dedicated mobile sensor platform. This can be a vehicle equipped with remote sensing devices such as cameras and/or LiDAR sensors as well as with hardware for direct georeferencing (i.e., an INS and its components). The data product generated this way is typically obtained through massive post-processing of the sensor data and consists of, for example, content for a geographic information system (GIS), 2D digital maps, or 3D city models.

Obviously, MODISSA with its data recording system described in Subsection~\ref{subsec:recording} and the direct georeferencing technique explained in Subsection~\ref{subsec:geo} can be referred to and used as a mobile mapping system. In this context, we work on the following research topics:
\begin{itemize}
\item Generation of 3D terrain models which can be used, for example, for LiDAR-based self-localization~\cite{gordon2016}.
\item Terrain navigability analysis~\cite{gehrung2019}.
\item 3D change detection in urban areas~\cite{gehrung2020}.
\item Semantic interpretation of large-scale 3D point clouds of urban areas~\cite{zhu2020}.
\end{itemize}
\subsubsection{The TUM-MLS-2016 benchmark dataset}
\label{subsubsec:tum-mls-2016}
\begin{figure}[htbp]
\centering
\includegraphics[width=\columnwidth,viewport=55 197 707 449,clip=true]{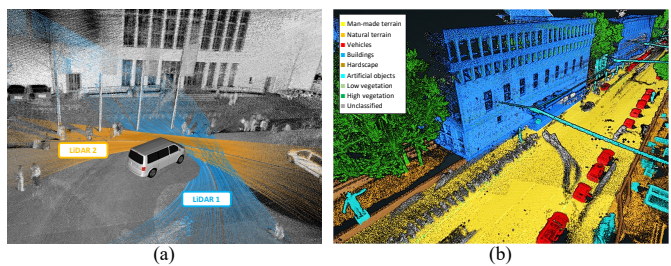}
\caption{Detailed views of the TUM-MLS-2016 data: (a) Data acquisition with two obliquely rotating LiDAR scanners, (b) different annotated object classes of the final point cloud.}
\label{fig:tum-mls}
\end{figure}

With regard to the last bullet point, this subsection provides some details of the data collection and data preparation that led to the ``TUM-MLS-2016'' benchmark dataset~\cite{zhu2020}. The relevant MLS data have been acquired in April 2016 using MODISSA. At the time of data collection in 2016, both Velodyne HDL-64E LiDAR scanners were positioned on wedges at a 25{\textdegree} angle to the horizontal, rotated outwards at a 45{\textdegree} angle. The orientation of the scanners can be seen in Figure~\ref{fig:detailed-views}a, top right, and the resulting scan pattern is shown in Figure~\ref{fig:tum-mls}a. Furthermore, each LiDAR scanner was configured to rotate at a frequency of 10~Hz and acquired approximately 130,000 range measurements (3D points) per rotation with distances up to 120~m. The LiDAR data were recorded synchronously with position and orientation data of the Applanix POS LV 520 INS, which were enhanced by GNSS correction data of the local GNSS reference station from the German SAPOS network. Thus, it was possible to perform quite accurate direct georeferencing of the LiDAR data and to aggregate all resulting 3D points in a common geocoordinate frame (ECEF).

The data acquisition took place in the area of the city campus of the Technical University of Munich (TUM). Within half an hour, MLS data in more than 17~thousand 360{\textdegree} 3D scans were acquired by each of the two LiDAR scanners as we drove MODISSA along the streets around the TUM city campus and its inner yard (cf. Figure~\ref{fig:tum-mls}a). The final point cloud consists of 1.7~billion 3D points in total, and covers an urban scenario of 70,000~m${^2}$ consisting of building facades, trees, bushes, parked vehicles, roads, meadows and so on. Each point has $x$, $y$, and~$z$~coordinates and carries the intensity $i$ of the laser reflectance. The handling of such amounts of 3D data can't be done straightforward, it requires an adequate and efficient data structure. Existing alternatives can be considered for the visualization, e.g., the open-source renderer ``Potree''~\cite{schuetz2015} which uses a hierarchical data structure in order to visualize the point cloud in real-time in a web browser (example: \href{https://s.fhg.de/vmls1}{s.fhg.de/vmls1}).

In the time-consuming manual annotation process, a subset (30\%) of the georeferenced points in the scene was manually labeled on a 20~cm voxel grid with eight semantic classes following the ETH standard~\cite{hackel2017} and one \textit{unclassified} class. An example showing the different classes can be seen in Figure~\ref{fig:tum-mls}b. In addition to the object classes, individual instances have also been annotated.

These MLS data with annotated ground truth are well suited for the development of methods for semantic scene interpretation, city modeling, but also for the investigation of real-time applications like object detection or LiDAR-SLAM. The complete dataset is made available to the scientific community under a Creative Commons License (\href{https://s.fhg.de/mls1}{s.fhg.de/mls1}). In combination with the follow-up dataset ``TUM-MLS-2018'', which is also available, methods for automatic 3D change detection can be developed and tested.
\subsubsection{Lessons learned (MODISSA and mobile mapping)}
\label{subsubsec:llmm}
On MODISSA, 360{\textdegree} 3D scanning LiDAR sensors are used which were not originally conceived for the purpose of mobile mapping. The measuring directions of the 64 individual laser rangefinders in the scanner head of a Velodyne HDL-64E are primarily oriented downward (2.0{\textdegree} up to 24.8{\textdegree} down). For use in the mobile mapping sector, e.g. for the mapping of facades, an unconventional tilted orientation of the sensors is therefore required, but this has proven to be quite successful in our measurement runs. The 360{\textdegree} 3D LiDAR scanners have the characteristic that they scan the vehicle environment overlapping at a rate of 10~Hz, which causes moving objects to leave a point trail in the accumulated 3D point cloud. Motion artifacts inevitably occur in mobile mapping of populated environments, but are particularly noticeable with this type of scanner. To remove artifacts caused by moving objects from the point clouds, filtering methods must be applied~\cite{gehrung2017}.

A successful feature is the consistent and accurate embedding of metadata in all data streams (time and location), which greatly simplifies offline sensor data fusion. However, the accumulation of data sequences also reveals offsets between multiple captures of the same location, which can be attributed to the inaccuracy of GNSS-based positioning. Even post-processing using correction data of a GNSS reference station cannot completely eliminate these errors, so loop closures and a bundle adjustment may still be necessary.

Attention must also be paid to handling the huge amounts of data that are created. Even just storing the data requires a careful selection of hardware (data links, data storage) that is suitable for the data rates that occur. By far the largest portion of MODISSA's data volume is caused by the video cameras (90\%), not by the LiDAR sensors. In a typical setting and using all sensors, currently about 4~TB of sensor data are generated within one hour.  
\subsection{Localization and observation of pedestrians in road traffic}
\label{subsec:persons}
Pedestrians are the most vulnerable road users, especially in an urban environment where pedestrians and vehicles have to share the traffic space. We use MODISSA for experiments in the field of automatic detection and observation of pedestrians in road traffic, with the aim of developing driver assistance systems or autonomous driving functions to ensure unconditional avoidance of accidents involving pedestrians.
In this respect, we believe that an active and passive multi-sensor approach offers the greatest reliability, given the differing advantages and disadvantages of applicable sensor technologies.

360{\textdegree} 3D scanning LiDAR is very efficient in capturing the 3D scene geometry, but its spatial resolution is usually too low to obtain fine details, such as the viewing direction of a detected person. In addition, LiDAR cannot capture color information of the scene. Both could greatly enhance the capabilities for automatic situational awareness, e.g., to assess the level of attention of nearby pedestrians. In our approach, we use the vehicle's LiDAR sensors for pedestrian detection and 3D tracking~\cite{borgmann2019, borgmann2020}. The determined positions of pedestrians are then used to select areas of interest in the images provided by the vehicle's panoramic cameras, which can then be further analyzed or highlighted to the driver.
\subsubsection{LiDAR-based detection of pedestrians}
\begin{figure}[htbp]
	\begin{center}
		\includegraphics[width=\columnwidth,viewport=27 174 741 466,clip=true]{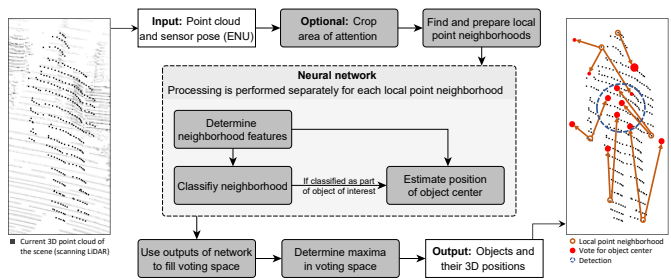}
		\caption{Procedural overview: LiDAR-based detection of relevant objects in the vehicle's vicinity.}
		\label{fig:pedestrian_detection_method_overview}
	\end{center}
\end{figure}

MODISSA provides us with consistent georeferenced MLS point clouds from multiple LiDAR sensors, taking the motion of the sensor platform into account, cf. Subsection~\ref{subsec:geo}. We have developed a method for LiDAR-based real-time detection of pedestrians, that takes these single point clouds as the main input, e.g., the 360{\textdegree} 3D scans of a rotating LiDAR scanner. The method is summarized in an overview in Figure~\ref{fig:pedestrian_detection_method_overview}.

In short, we combine two different machine learning methods to achieve object detection, classification and localization within 3D point clouds:
\begin{itemize}
\item \textit{Analysis of local point neighborhoods within the point cloud}. This is intended to detect parts of relevant objects, e.g., body parts of persons. We accomplish this task via a neural network that is inspired by ``PointNet''~\cite{qi2017}.
\item \textit{Detection of relevant objects with a 3D voting process}. Here, each previously found object part casts votes for possible center positions of the parent object. The objects are revealed as the emerging clusters (maxima) in the voting space. This step is inspired by the ``Implicit Shape Model''~\cite{leibe2008}.
\end{itemize}
In preparation of the processing step corresponding to the first bullet point, the input point cloud is first cropped to an area of attention: the approximate location of the ground surface relative to the sensor is known, allowing only points above it and up to a height of two meters to be considered relevant. \textit{Local point neighborhoods} are then analyzed, where each such neighborhood is generated by a center point and its surrounding points within a certain radius, which is chosen to be approximately 7~times the average point-to-point distance in the scene. The neural network that processes these local point neighborhoods is inspired by ``PointNet'' and can cope with unstructured 3D points, hence no grid discretization is required.
The neural network is composed of three parts, see Figure~\ref{fig:pedestrian_detection_method_overview} and~\cite{borgmann2019}:
\begin{enumerate}
	\item Extraction of descriptive neighborhood features.
	\label{itm:nn1}
	\item Classification of the neighborhood as part of an object of a certain type.
	\label{itm:nn2}
	\item Regression to estimate the center of the parent object, if relevant.
	\label{itm:nn3}
\end{enumerate}
Component~\ref{itm:nn1} of the neural network is a \textit{multi-layer perceptron} with four layers followed by a max-pooling layer. It is trained to discover features of local point neighborhoods that are characteristic for parts of certain object classes, e.g., human body parts. The training itself needs manually annotated point clouds, however, only a comparatively low amount of these labeled training data is required as they split combinatorically into many labeled local point neighborhoods. Component~\ref{itm:nn2} and Component~\ref{itm:nn3} both constitute subsequent multi-layer perceptrons to classify the parent object based on the local feature vector with a certain degree of confidence, and to estimate this object's center position in a 3D voting space. In a multi-sensor system, the voting space can even accumulate the votes originating from multiple LiDAR sensors.

Finally, each cluster (maximum) that is found in the voting space represents an object and yields the respective result: object class, object position, and 3D bounding box containing all detected parts of the object. Note that although pedestrians are the most relevant object class for this subsection, we consider other object classes as well (bicyclists, vehicles, etc.).

\begin{figure}[htbp]
	\begin{center}
		\includegraphics[width=\columnwidth]{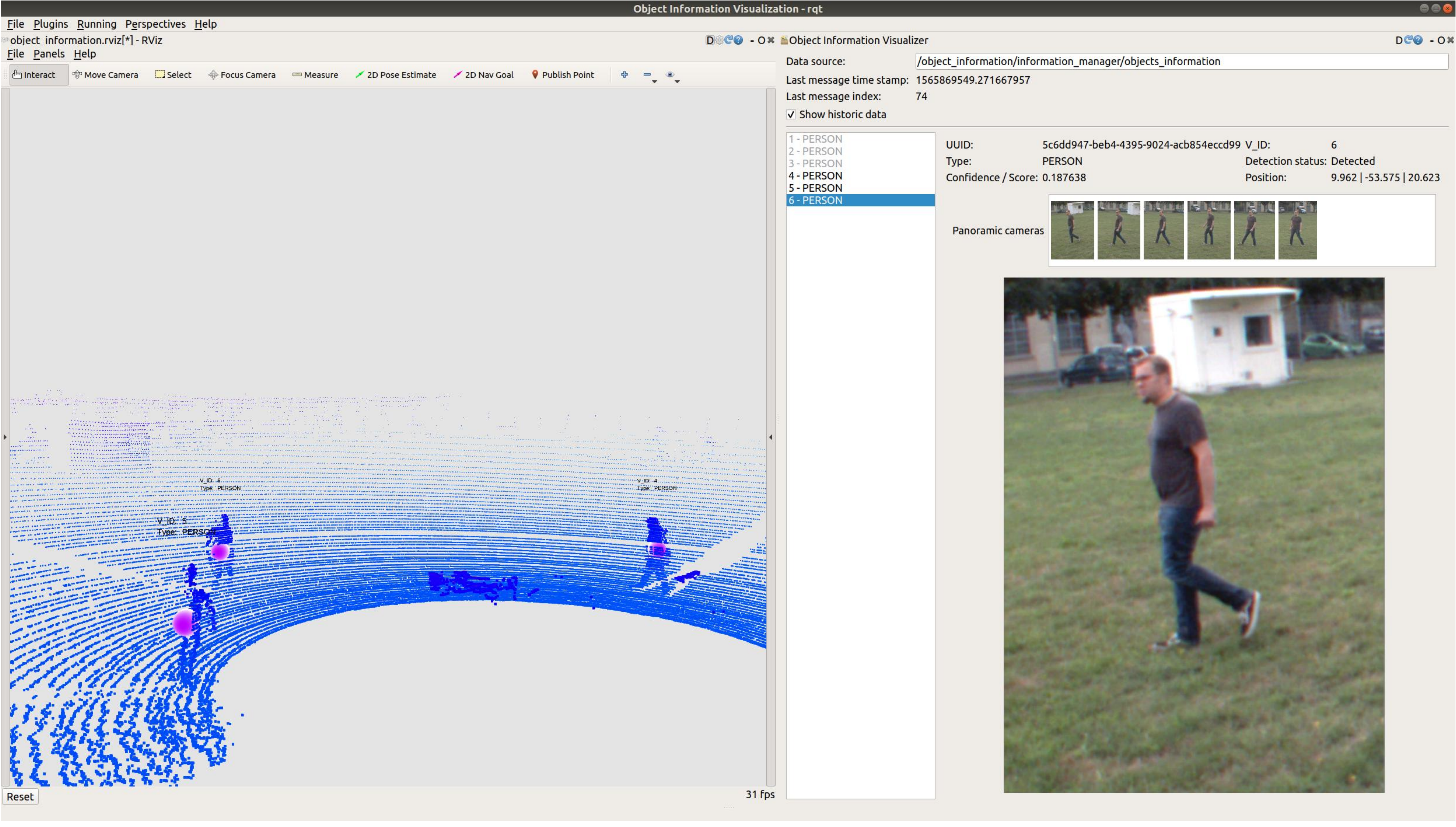}
		\caption{MODISSA's on-board display shows real-time results of pedestrian detection and observation in the vicinity of the vehicle.}
		\label{fig:pedestrian_detection_example}
	\end{center}
\end{figure}
\subsubsection{Observation of pedestrians and handoff to other sensors}
LiDAR-based detection is complemented by a method for tracking pedestrians based on a Kalman filter. By keeping track of detected pedestrians, it is possible to observe and automatically analyze their behavior, e.g., to avoid accidents. Furthermore, this approach allows to keep track of temporarily occluded pedestrians and simplifies re-detection of the same pedestrians in a scan sequence~\cite{borgmann2020}.

MODISSA's real-time processing system (see Subsection~\ref{subsec:setup_realtime}) handles time-synchronous sensor data, and ROS maintains the complete transformation chain between fields-of-view of all sensors (extrinsic and intrinsic). This allows to project information regarding detected pedestrians from the LiDAR scans into corresponding fields-of-view, e.g., of the panoramic cameras. Since this can be done in real-time, simultaneous analysis of the pedestrians in the LiDAR and visual sensor data is made possible. We currently still work on this research topic, but we can provide a first insight: Figure~\ref{fig:pedestrian_detection_example} shows a typical output of the real-time processing in ROS, displayed on MODISSA's on-board monitor. The left side contains a 3D viewer with the results of LiDAR-based person detection. Image sections following the 3D track of a detected and selected pedestrian are shown on the right with additional metadata.
\subsubsection{Lessons learned (MODISSA and pedestrian detection)}
\label{subsubsec:llped}
Overall, our approach for LiDAR-based pedestrian detection performs reasonably well, as long as the local point density is not too low. Regarding the Velodyne HDL-64E LiDAR sensors we currently still use with MODISSA, this is the case up to distances of 22~m. A detailed quantitative evaluation can be found in~\cite{borgmann2019, borgmann2020}. While detection of obstacles and movements is still possible even at longer distances, classifying the specific object type becomes increasingly difficult. The Velodyne VLP-16 offers an even lower resolution, so that it is actually ruled out for long-range classification tasks.

Remarkably, we notice a decrease in classification performance when the orientation of the LiDAR sensors is changed (see Figure~\ref{fig:detailed-views}a) in comparison to that of the training data recordings. A possible reason could be the change in direction-dependent angular resolutions, for which the neural network does not generalize well. For the Velodyne HDL-64E rotating at 10~Hz, the horizontal resolution is 0.18{\textdegree}, whereas its vertical resolution is 0.4{\textdegree}. We will soon replace MODISSA's front LiDAR sensors with two Ouster OS2-128 which can have a more homogeneous resolution in both directions, so we will be able to analyze this effect further.

Mechanically operating LiDAR scanners unavoidably lead to distortion effects on moving objects in the scene, and combining LiDAR data from multiple scanners consequently leads to mismatches. The same effect is found in image data acquired with rolling shutter cameras. In this analogy, a ``global shutter'' would be preferable, i.e., all 3D points of a single point cloud unit should be acquired at exactly the same time. Flash LiDAR is one technology that can do this, but it is currently not quite adequate for capturing the 360{\textdegree} surroundings. However, mechanical LiDAR scanners may be superseded by another technology with this capability in the near future. 

Our experiments with MODISSA in the context of this application are substantially supported by the ROS software environment. It turned out to be best practice to include all implemented algorithms in a software library. This way, the library can be linked against a ROS program on MODISSA for online processing, but it can also be linked against a user interface for offline processing of recorded data. This avoids duplicate work and also simplifies debugging.
\subsection{Protection of vehicles against UAV attacks}
\label{subsec:uavs}
\begin{figure}[htbp]
\centering
\includegraphics[width=\columnwidth,viewport=41 162 731 498,clip=true]{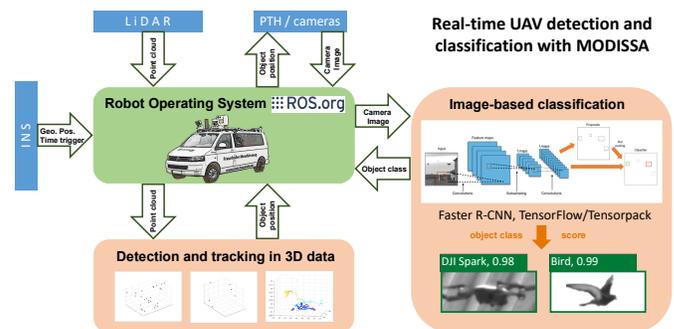}
\caption{Schematic overview of real-time UAV detection, tracking, and classification. ROS is used to implement the interaction of sensor data acquisition and sensor control with methods for real-time data processing.}
\label{fig:ROSstruct01}
\end{figure}

Due to the availability and ease of use of small drones (UAVs, unmanned flying vehicles), the number of reported dangerous incidents caused by them, both with malicious intent and accidentally, is increasing. To prevent such incidents in the future, it is necessary to be able to detect approaching UAVs. Entities to be protected against UAVs are often stationary, such as airports or industrial facilities. There is also a growing need for security and protection against UAVs at high-level events such as the annual G7 summits. Several vendors already offer technical solutions which promise to detect all approaching UAVs in order to initiate appropriate countermeasures (examples: Dedrone DroneTracker, APsystems SKYctrl). Typically, such solutions are operated stationary and most providers apply a combination of several sensors. Usually cameras in the visual and IR spectrum are combined with radar, radio, and acoustic sensors. Nevertheless, especially with regard to the future topic of autonomous driving as well as in the security and military sector, moving vehicles also require protection against collisions with flying objects. It would be preferable if the sensors already installed on the vehicle could be used for this additional task. In using MODISSA as the experimental system, we developed an approach for mobile detection and classification of flying objects in the vicinity of the vehicle. The complete chain of multi-sensorial data acquisition, data pre-processing, data handling, object detection, sensor alignment, and object classification is performed in MODISSA in real-time (see Figure~\ref{fig:ROSstruct01}). Although there are analogies to the person detection described in the previous subsection, UAVs require a different treatment as they are too small to be classified based on their extent in LiDAR point clouds.

\subsubsection{Detection of UAVs in 360{\textdegree} 3D LiDAR scans}
In the first step, we apply a geometric model-based algorithm to detect isolated 3D objects of suspicious size in the point clouds measured by the 360{\textdegree} 3D scanning LiDAR sensors~\cite{hammer2018}. Since LiDAR is an active sensing technology, it is independent of the scene illumination, therefore detection of UAVs is possible even in low-light conditions and at night. Moreover, the exact 3D position of detected objects is determined automatically: during direct georeferencing (see Subsection~\ref{subsec:geo}), all involved coordinate systems are accurately known, and the 3D position of detected objects can be transferred to the global geocoordinate system.

\subsubsection{3D tracking of detected objects}
\begin{figure}[htbp]
\centering
\includegraphics[width=\columnwidth,viewport=21 177 719 584,clip=true]{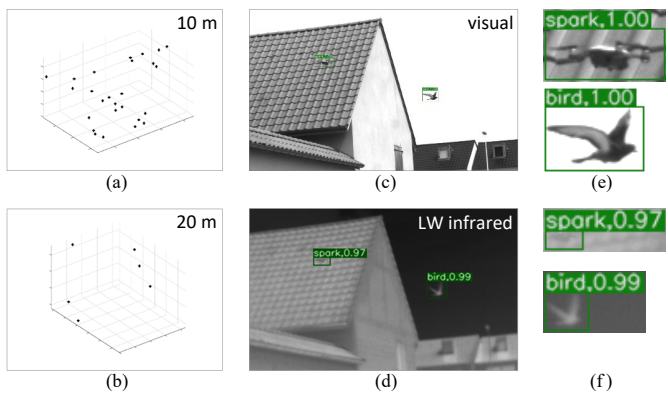}
\caption{(a) 3D point cloud representing a UAV in 10~m distance, (b) in 20~m distance. (c) Visual image with results of object classification. (d) LWIR image with results of object classification. (e) Details of c. (f) Details of d.}
\label{fig:UAV_pcd01}
\end{figure}

The scan pattern of 360{\textdegree} 3D scanning LiDAR sensors can cover the complete vicinity of the vehicle. Unfortunately, due to the low spatial resolution of such sensors, a detected object is typically captured by only a few measurement points, depending on the respective detection distance (as shown in Figures~\ref{fig:UAV_pcd01}a and~\ref{fig:UAV_pcd01}b).
With such sparse point clouds, classification of the detected objects in the 3D data is not very promising. To reduce the rate of false alarms, the detected objects can be tracked in successive 360{\textdegree} 3D scans. Detections of non-moving objects are discarded as they may be artifacts in the data, or at least they do not pose an immediate threat. For detected objects that significantly change their position in 3D space, their trajectory can be analyzed and thus their future position can be predicted.

\subsubsection{Image-based classification ``drone versus bird''}
Different flying objects of the expected size and with the expected motion pattern are conceivable. Not only typical kinds of mini and micro UAVs, but also birds often meet these criteria. Distinguishing between UAVs and harmless birds is the minimum expectation to the following classification step, but also an exact identification of the UAV type is desirable. This is especially useful if specific countermeasures are intended for individual UAV types, e.g., by jamming their radio control signal. At present, such object classification tasks are most commonly realized with machine learning methods, for example convolutional neural networks (CNNs). There exist approaches for point-cloud-based machine learning (see Subsection~\ref{subsec:persons}), but these are not useful for very sparse point clouds.

More promising is an image-based classification based on high-resolution camera images. Assuming multi-sensor equipment such as MODISSA's on future cars, cameras can be pointed at the 3D position of the detected object in a timely manner and images can be acquired, both visual and thermal infrared (LWIR). 
In preparation of the image-based classification step, we trained a CNN model (Faster R-CNN~\cite{ren2017}, implemented in a TensorFlow/Tensorpack framework) using about 15,000 manually annotated visual grayscale images that contained eight different UAV types and several bird species~\cite{hammer2020}. Using this neural network, we achieved good classification results for daylight scenes (Figure~\ref{fig:UAV_pcd01}c). For object classification in scenes with poor illumination and especially at night, when object detection by LiDAR is still feasible, we trained the same CNN model with LWIR images. The LWIR-based results were found to lag behind the performance of the visual camera under daylight conditions (Figure~\ref{fig:UAV_pcd01}d), but as expected, LWIR imaging enables classifying objects even in low-light scenarios. Regarding 24/7 applicability, we found the best results with a combination of the LiDAR-based object detection followed by a combined visual and LWIR image-based classification, achieving a precision of 96.1\% and a recall of 96.5\% in our experiments. Details of the evaluation of UAV detection and classification performance can be found in our two previous papers referenced in this subsection~\cite{hammer2018, hammer2020}.
\subsubsection{Lessons learned (MODISSA and UAV detection)}
\label{subsubsec:lluav}
We found that the detection of micro UAVs in LiDAR point clouds is feasible, even if LiDAR sensors are used that were not built for this purpose. However, the range of detection is highly limited by the angular resolution of the LiDAR sensor(s). We will soon be able to investigate this even more quantitatively, as replacement of MODISSA's Velodyne HDL-64E (vertical resolution 0.4{\textdegree}) LiDAR scanners with more recent Ouster OS2-128 (vertical resolution 0.2{\textdegree}) is currently in progress. Another desirable technical improvement of LiDAR sensor technology with regard to its use for monitoring in all spatial directions would be a significant increase in the vertical field-of-view, which for the Velodyne HDL-64E is limited to 27{\textdegree}, and in the horizontal coverage or 360{\textdegree} scanning performance. In this regard, we expect further advances even in automotive LiDAR technology, and new products from sensor manufacturers in response.

The comparatively slow mechanical alignment of the cameras mounted on the PTH with the detected flying object is a bottleneck for real-time UAV identification. This problem could be solved by using the vehicle-encompassing panoramic cameras, especially if such were also available for the IR spectrum. Enhancements are currently being made to MODISSA's sensor equipment for this purpose as well.

The ROS software environment provides a workbench for real-time UAV detection and offers flexibility, but also requires ongoing maintenance. This is not a problem for prototyping and process development in academia, but a dedicated real-time environment and implementation will be required for an operational system.
\section{Discussion and conclusions}
\label{sec:conclusions}
This paper describes the setup and selected applications of MODISSA as a mobile laboratory of our research group in the Department \textit{Object Recognition} (OBJ) at Fraunhofer~IOSB. By providing our empirical findings, we aim to help other groups who want to build a similar experimental system. Since this is the main goal of the paper and since there were many different aspects to cover, we have detailed our specific lessons learned in the respective subsections throughout the paper. This was done as a partial replacement for a separate discussion and summary in a single concluding section. For this reason, our sensor- and application-specific experiences related to
\begin{itemize}
\item \textit{extrinsic system calibration} (\ref{subsec:geo}),
\item \textit{data protection and anonymization} (\ref{subsec:anonym}),
\item \textit{mobile mapping} (\ref{subsubsec:llmm}), 
\item \textit{real-time object recognition and tracking} (\ref{subsubsec:llped} and \ref{subsubsec:lluav}),
\end{itemize}
are not repeated a second time. We wrap up the paper with a more general discussion and reflection on MODISSA, and with some additional thoughts beyond what has already been said in the paper.

MODISSA enables research in the contexts of area-wide multi-sensorial data acquisition and direct analysis of the sensor data for vehicle-related applications. In our selection of \textit{sensor technologies} (\ref{subsec:installation}) and \textit{data processing capabilities} (\ref{subsec:recording} and~\ref{subsec:setup_realtime}), we deliberately choose a balance between the vehicle's suitability for research purposes in connection with real-time applications~\ref{itm:rw1A} and also for data collection in surveying and mapping~\ref{itm:rw1B}. One may ask what trade-off we made to satisfy both use cases, and whether this is the kind of configuration everyone should follow.

To start with the second part of the question, the answer is probably ``it depends''. We made compromises that would not be made when developing a dedicated system. For example, vehicle-based UAV detection (\ref{subsec:uavs}): If this task were the sole basis for sensor selection, it would certainly differ from that of MODISSA. We can give two arguments why the configuration of MODISSA is nonetheless reasonable:
\begin{itemize}
\item \textit{Flexibility}:
As a research group, we need to be broad and flexible to work on a whole range of current and future research projects. Our main research focus is on the application-oriented development of new methods for sensor data processing, which is why we can compromise on sensor equipment and do not always require the latest sensors. Of course, we pay attention to current advances in the respective sensor technologies and make sure that our methods will be applicable on future operational systems, even though they were created with an experimental system that tends to be suboptimal for the respective application.
\item \textit{Multi-purpose sensors}:
We expect that almost all future cars will be equipped with optical sensors for dedicated purposes, e.g., to enable autonomous driving or driver assistance functions. Once this is achieved, it would be obvious to use these omnipresent sensors for other purposes as well, e.g., for mapping, change detection, or security tasks. Apart from ethical questions this raises, there is still the technical challenge of performing a task with sensors that were not primarily designed for that purpose. We consider this to be an interesting field of research.
\end{itemize}

The MODISSA testbed has already undergone several upgrades in recent years. We will further evolve it, and MODISSA will continue to be used for detailed experiments and tests within a number of projects. Furthermore, we are highly interested in studying applications in which sensor-equipped vehicles automatically exchange information. In the near future, we may therefore build an additional MODISSA-2, based on our experiences so far.
\section*{Acknowledgments}
The authors would like to thank Ann-Kristin Grosselfinger and David M\"unch from Fraunhofer~IOSB for their work related to the anonymization of MODISSA's sensor data. We also thank Jingwei Zhu, Zhenghao Sun, and Yusheng Xu from the Technical University of Munich for their efforts in annotating the TUM-MLS-2016 and TUM-MLS-2018 datasets. Parts of this work were carried out within the framework of the Fraunhofer Cluster of Excellence ``Cognitive Internet Technologies''. The sensor data used to evaluate the UAV detection were obtained during a measurement campaign of the NATO group SET-260 (``Assessment of EO/IR Technologies for Detection of Small UAVs in an Urban Environment'') at the CENZUB facility, Sissonne, France.

\section*{Disclosures}

The authors declare no conflicts of interest.

\bibliography{iosb-bib-modissa}

\end{document}